\documentclass[letterpaper]{article} 
\usepackage{aaai25}  
\usepackage{times}  
\usepackage{helvet}  
\usepackage{courier}  
\usepackage[hyphens]{url}  
\usepackage{graphicx} 
\usepackage{natbib}  
\usepackage{caption} 
\usepackage{algorithm}
\usepackage{algorithmicx}
\usepackage{algpseudocode}

\usepackage{newfloat}
\usepackage{listings}
\usepackage{arydshln}
\usepackage{color}
\usepackage{graphicx}
\usepackage{amsmath,bm}
\usepackage{amssymb}
\usepackage{array}
\usepackage{comment}
\usepackage{booktabs}
\usepackage{subfigure}
\usepackage{caption}
\usepackage{colortbl} 
\usepackage{multirow}
\usepackage{makecell}
\usepackage{pifont}
\usepackage{tabularray}
\usepackage{paralist}
\usepackage{adjustbox}
\usepackage{rotating}
\usepackage{enumerate}
\usepackage{tcolorbox}
\usepackage{fontawesome5}
\usepackage{tabularx}
\usepackage{booktabs}       
\usepackage{soul}
\usepackage{bbding}
\usepackage{comment}
\usepackage{booktabs}
\usepackage{dashrule} 
\usepackage{xcolor}
\usepackage{xspace}

\DeclareCaptionStyle{ruled}{labelfont=normalfont,labelsep=colon,strut=off} 
\floatstyle{ruled}
\newfloat{listing}{tb}{lst}{}
\floatname{listing}{Listing}
\definecolor{myblue}{RGB}{52,218,247}
\definecolor{myred}{RGB}{255,90,90}
\definecolor{mypink}{RGB}{239,43,159}
\definecolor{myupdate}{RGB}{254,243,222}
\definecolor{myfrozen}{RGB}{237,255,255}
\definecolor{ired}{RGB}{229,72,72}
\definecolor{igreen}{RGB}{80,219,144}
\definecolor{nmblue}{RGB}{216,234,247}
\definecolor{lightblue}{RGB}{211,227,253}
\definecolor{lightgrey}{RGB}{201,203,209}
\definecolor{linkred}{RGB}{255,0,49}
\definecolor{textred}{RGB}{255,242,234}
\definecolor{imageblue}{RGB}{224,250,255}
\definecolor{videogreen}{RGB}{214,250,232}
\definecolor{nmgray}{RGB}{229,229,229}
\definecolor{boxgrey}{RGB}{154,186,211}
\definecolor{lightgreen}{RGB}{208,255,233}
\definecolor{lightred}{RGB}{251,217,226}

\definecolor{defgreenDark}{RGB}{0,176,80}
\definecolor{defgreenLight}{RGB}{208,255,233}
\definecolor{defredLight}{RGB}{251,217,226}

\newtcolorbox{mybox}[2][]{
width=\textwidth,
colback = nmgray!75!white, 
colframe = nmgray!75!white, 
boxsep=0pt,left=9pt,right=10pt,top=0pt,bottom=0pt,
fontupper=\linespread{0.9}\selectfont,
title=#2,#1}

\newtcolorbox{instructionbox}[2][]{
  colback=boxgrey!5!white, 
  colframe=boxgrey!75!black, 
  fonttitle=\bfseries, 
  title=#2,#1
  boxsep=1pt, 
  left=1pt, 
  right=1pt, 
  top=1pt, 
  bottom=1pt 
}

\newcommand{\SG}{\raisebox{-4pt}{\includegraphics[width=1.5em]{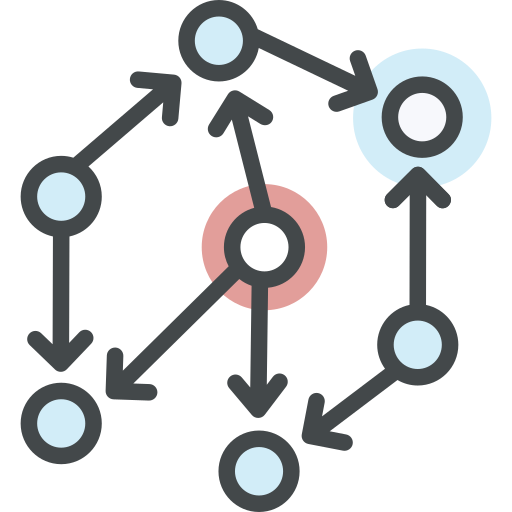}}\xspace\xspace}
\newcommand{\image}{\raisebox{-3pt}{\includegraphics[width=1.5em]{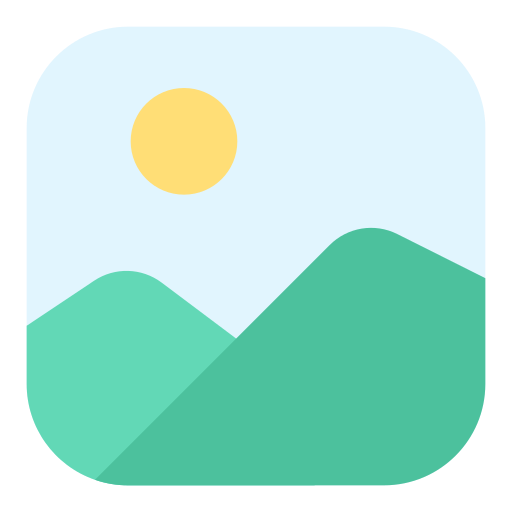}}\xspace\xspace}

\makeatletter
\newcommand\figcaption{\def\@captype{figure}\caption}
\newcommand\tabcaption{\def\@captype{table}\caption}
\makeatother

\title{Combating Multimodal LLM Hallucination via Bottom-Up Holistic Reasoning}
\author {
    Shengqiong Wu\textsuperscript{\rm 1},
    Hao Fei\textsuperscript{\rm 1}\thanks{Corresponding author.}, 
    Liangming Pan\textsuperscript{\rm 2}, 
    William Yang Wang\textsuperscript{\rm 3} \\
    Shuicheng Yan\textsuperscript{\rm 4,5}, 
    Tat-Seng Chua\textsuperscript{\rm 1}
}
\affiliations {
    \textsuperscript{\rm 1}National University of Singapore, Singapore \\
    \textsuperscript{\rm 2}University of Arizona, USA\\
     \textsuperscript{\rm 3}University of California, Santa Barbara, USA \\
    \textsuperscript{\rm 4}Skywork AI, Singapore \\
    \textsuperscript{\rm 5}Nanyang Technological University, Singapore\\
    swu@u.nus.edu, \{haofei37, dcscts\}@nus.edu.sg, liangmingpan@arizona.edu\\
    william@cs.ucsb.edu, shuicheng.yan@gmail.com
}

\begin{document}

\maketitle

\begin{abstract}
Recent advancements in multimodal large language models (MLLMs) have shown unprecedented capabilities in advancing various vision-language tasks.
However, MLLMs face significant challenges with hallucinations, and misleading outputs that do not align with the input data. 
While existing efforts are paid to combat MLLM hallucinations, several pivotal challenges are still unsolved.
First, while current approaches aggressively focus on addressing errors at the perception level, another important type at the cognition level requiring factual commonsense can be overlooked.
In addition, existing methods might fall short in finding a more effective way to represent visual input, which is yet a key bottleneck that triggers visual hallucinations.
Moreover, MLLMs can frequently be misled by faulty textual inputs and cause hallucinations, while unfortunately, this type of issue has long been overlooked by existing studies.
Inspired by human intuition in handling hallucinations, this paper introduces a novel bottom-up reasoning framework.
Our framework systematically addresses potential issues in both visual and textual inputs by verifying and integrating perception-level information with cognition-level commonsense knowledge, ensuring more reliable outputs.
Extensive experiments demonstrate significant improvements in multiple hallucination benchmarks after integrating MLLMs with the proposed framework.
In-depth analyses reveal the great potential of our methods in addressing perception- and cognition-level hallucinations.

\end{abstract}

\section{Introduction}
\label{introduction}

\begin{figure}[!t]
\centering
\includegraphics[width=1.0\columnwidth]{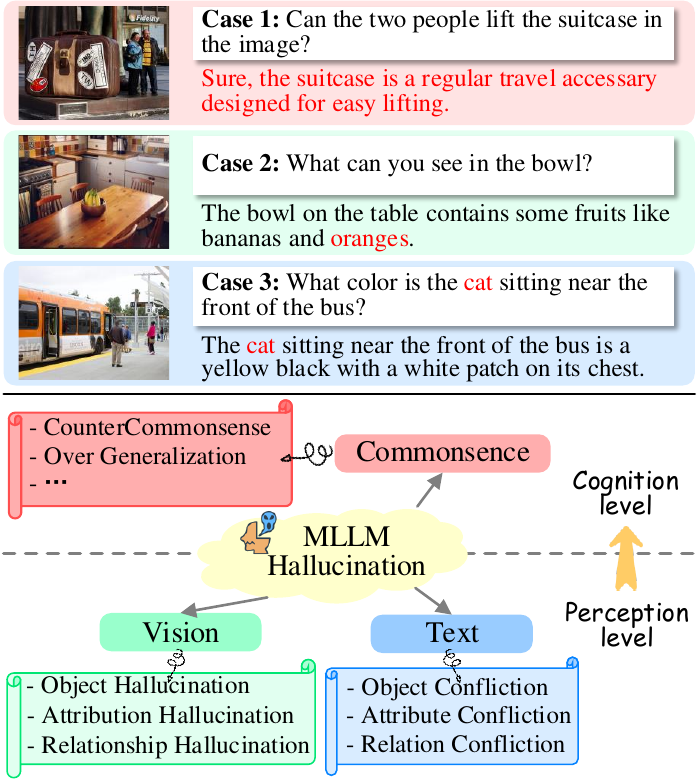}
\caption{
On the top, we illustrate three hallucination cases: overgeneralization (case 1), vision object hallucination (case 2), and text object conflict  (case 3), where hallucinations are marked in \textcolor{red}{red}. On the bottom, we categorize hallucinations into three types: vision, text, and commonsense, ranging from perception to cognition levels.
}
\vspace{-6mm}
\label{fig:intro}
\end{figure}

Recent advancements in MLLMs \cite{abs-2310-03744,abs-2311-03079,abs-2310-09478} have demonstrated impressive abilities in understanding the semantics of multimodal data and achieving promising results in a variety of tasks such as visual question answering \citep[VQA; ][]{gurari2018vizwiz,00010BT0GZ18}, multimodal dialogue \cite{abs-2303-04671,wang2024modaverse,wu2023next}, image captioning \cite{MilewskiMC20,abs-2310-03744,ji2021improving}, and retrieval \cite{fang2024not,LiWQNLC24,fang2024fewer}. 
However, much like the hallucination issues observed in traditional LLMs due to their generative nature, MLLMs also suffer from hallucinations, leading to factual discrepancies between the model's response and factual reality \cite{abs-2401-06209,abs-2309-05217}. 
This significantly undermines their practical value. 
Various types of hallucinations have been identified, prompting many research efforts to address this problem. 
Existing approaches include fine-tuning MLLMs on more robust instruction datasets \cite{WangHLLL24,liu2024mitigating}, proposing vision-aware inference interventions \cite{abs-2403-14003,deng2024seeing,fang2023you}, and employing a post-correction method where results are generated \cite{abs-2402-11622,abs-2310-16045}, then checked and corrected if necessary. 
Despite these efforts, several critical issues remain unresolved in mitigating hallucinations, especially after deeply revisiting the root-cause of hallucinations from MLLM.

\textbf{Firstly}, a fundamental issue that results in hallucinations in MLLMs is rooted in their insufficient understanding of visual input, which stems from the fact that language-based LLMs do not perceive visual content with the same depth as they understand language. 
Thus, there is a crucial lack of effective visual representations that assist MLLMs' comprehension of visual data \cite{fang2025rethinking,fang2022multi,li2022fine,li2022compositional}.
We propose adopting a more structured visual scene representation method, namely Scene Graph (SG), which effectively captures the semantic structure of key objects, attributes, and relationships \cite{KrishnaZGJHKCKL17,Wu00BC23,abs-2311-17076}. 
This representation naturally aids in addressing the three crucial types of visual hallucinations: object, attribute, and relationship hallucinations \cite{abs-2407-03314}. 
It is important to note that while current de-hallucination efforts tend to focus on object and attribute hallucinations, typically by employing object detection and phrase grounding tools, often overlooking relationship hallucinations. 
The introduction of SG representation is expected to resolve these types of issues effectively.

\textbf{Secondly}, existing work often overlooks the fact that input text itself can also be prone to hallucinations. 
For example, as shown in Figure \ref{fig:intro}, the input text question, ``\texttt{What color is the cat sitting near the front of the bus?}'' contradicts the image as the input image does not actually present a cat. 
Such discrepancies may mislead MLLMs, resulting in erroneous outputs that perpetuate hallucinations.
Alarmingly, our statistics show that up to 47.8\% of user input texts contain such discrepancies (detailed in Appendix \S C.3), severely impacting task performance.

\textbf{Lastly}, upon reevaluating existing methods to address multimodal hallucinations, we observe that most efforts focus on mitigating perception-level hallucinations, such as through object and attribute grounding, but neglect cognition-level checks. 
Notably, our analysis indicates that issues reliant on cognition-level commonsense knowledge constitute approximately 51\% of the problems, particularly in complex input queries.
Drawing inspiration from human intuition, we argue that addressing hallucinations should be a holistic reasoning process that incorporates a bottom-up approach from perception to cognition grounding, where perception addresses low-level content awareness and cognition tackles the factuality of commonsense knowledge.

To address the aforementioned challenges, we propose a novel bottom-up reasoning framework for MLLM de-hallucination, from low-level perception grounding to high-level cognition grounding. 
Inspired by the chain-of-thought \citep[CoT; ][]{abs-2311-17076,YaoZYDSN023,fei2024video,fei-etal-2023-reasoning}, our framework decomposes the raw visual questions into smaller reasoning subprocesses. 
These subprocesses progress from high-fidelity perception of target content in both input images and texts to accurately responding with validated cognition-level knowledge.
Specifically, our approach can be broken down into six reasoning modules.
\textcircled{\small{\textbf{1}}}: We guide the MLLM to focus on the visual area most relevant to the user's question, prompting it to generate a partial scene graph. This step ensures the capture of complete visual information, including {objects}, {attributes}, and {relationships}), essential for answering the question.
\textcircled{\small{\textbf{2}}}: Using external tools, we verify and correct the focused visual content of the partial scene graph representation, ensuring the accuracy of the perceived visual content and preventing hallucinations at the perception level.
\textcircled{\small{\textbf{3}}}: Based on the high-faithfulness visual perception, we further verify and rectify any discrepancies in the input question that may conflict with the visual content, ensuring accuracy and consistency between the input visuals and text.
\textcircled{\small{\textbf{4}}}: 
After confirming the accuracy of both the visual content and the textual question, MLLMs should be able to answer perception-level questions. 
However, this may not always suffice for cognition-level questions
In such cases, the need for additional cognition-level knowledge arises, prompting the MLLM to generate the necessary commonsense claims to answer the question.
\textcircled{\small{\textbf{5}}}: We further verify MLLM-induced commonsense claims against an external knowledge base.
\textcircled{\small{\textbf{6}}}: Integrating all verified perception-level information and cognition-level commonsense, we direct the MLLM to produce the final answer.

We conduct extensive experiments on six benchmarks, demonstrating that the existing MLLMs equipped with our proposed method show significant improvement in mitigating hallucination.
In-depth analyses and visualizations show that our method helps decrease conflicts in input questions, thereby reducing erroneous outputs.
Overall, our contributions can be summarized into four aspects:
\setdefaultleftmargin{2.0em}{0.2em}{1.87em}{2.7em}{1em}{1em}
\begin{compactitem}
    \item Drawing inspiration from human reasoning, we propose a novel holistic bottom-up reasoning framework for MLLM de-hallucination, spanning from perception to cognition.
    
    \item Our framework innovatively utilizes scene graph representations for visual content during the de-hallucination process of MLLMs. 
    
    \item  We are the first to highlight the often-overlooked issue of hallucinations originating from input text questions, which significantly contribute to hallucinations in user interactions. To address this, we introduce a reconsideration mechanism designed to reduce conflicts between the input queries and visual content, thereby avoiding misleading responses.

    \item Our proposed framework effectively mitigates various types of hallucination within MLLMs, demonstrating its potential and comprehensiveness.
\end{compactitem}

\vspace{-2mm}
\section{Related Work}
\label{related_work}

Despite the unprecedented capabilities of LLMs, they still exhibit errors on certain NLP tasks, aka., hallucination, due to their generative nature \cite{abs-2402-00253}. 
When expanded to MLLMs \cite{wu2024towards,fei2024vitron,abs-2306-02858}, such an issue persists, characterized by generated text responses that don't align with corresponding visual content \cite{abs-2310-16045}. 
Research \cite{abs-2311-13614,abs-2401-06209,abs-2310-00754} categorize hallucinations of MLLM in three types: \emph{object hallucination}, \emph{attribute hallucination}, and \emph{relation hallucination}. 
To address these, various solutions have been proposed \cite{abs-2310-00754,abs-2310-16045,liu2024mitigating}. 
Some suggest refining MLLM with cleaner, more accurate training data \cite{WangHLLL24}, while others advocate for model intervention during inference \cite{abs-2403-14003,deng2024seeing}, or directly correct model outputs \cite{abs-2402-11622,abs-2310-16045}.
In this work, we thoroughly rethink the triggers of MLLM hallucination, where certain key aspects that existing works have not fully considered.

On the one hand, MLLMs \cite{li2022fine} often lack a detailed and accurate understanding of visual images, leading to erroneous outputs. 
To address this, we propose using scene graph (SG) representations to enhance image comprehension. 
SGs \cite{abs-2403-12033,CongYR23,fei2024enhancing}, highly structured image representations, that precisely capture the semantic meanings of objects, attributes, and their relationships, can intuitively help mitigate all the above three hallucination types by enabling fine-grained and controllable checks of visual faithfulness.
On the other hand, aside from the understanding of vision itself, many hallucinations can stem from user textual queries that contain inconsistencies with the visual inputs, misleading MLLM outputs. 
Unfortunately, this cause has been largely overlooked in prior studies \cite{abs-2402-11622,abs-2310-16045,abs-2403-14003,li2022compositional}.

\begin{figure*}[!t]
\centering
\includegraphics[width=0.92\textwidth]{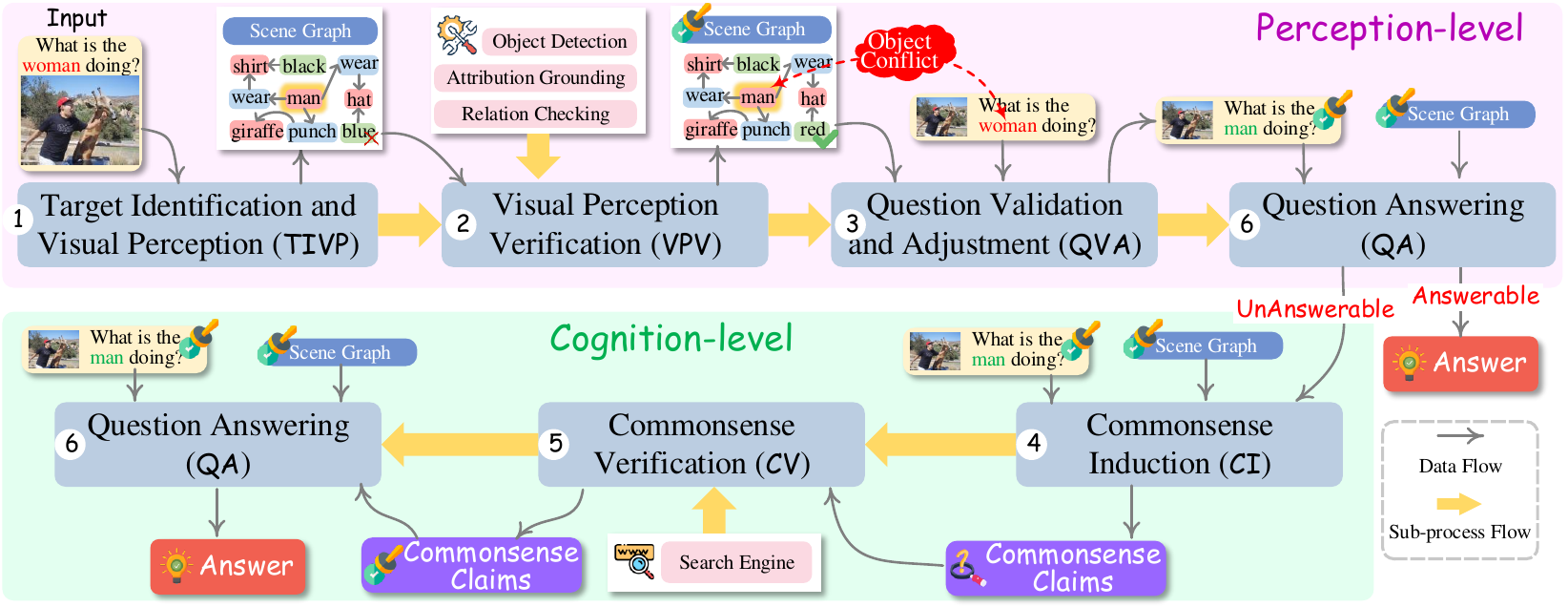}
\vspace{-2mm}
\caption{
Illustration of the overall framework of \textsc{Dehall}, consisting of six reasoning modules from perception to cognition. 
}
\vspace{-5mm}
\label{fig:framework}
\end{figure*}

Last but not least, most research focuses solely on content faithfulness at the recognition level \cite{abs-2311-07397,LiDZWZW23,abs-2403-11116,abs-2309-05217,ji2022knowing}, neglecting the factuality checks at the cognitive aspect necessary to prevent hallucinations. 
Yet this is a critical source for the occurrence of hallucinations, as MLLMs often provide counterfactual responses due to a lack of commonsense \cite{abs-2403-11116}.
Thus, we propose a holistic reasoning framework, inspired by human experts who employ a strict `\emph{from-recognition-to-cognition}' process. 
We first ensure that all input content (including images and texts) is correctly recognized at the recognition level. 
Upon building a correct cognition of the inputs, further deep reasoning about the factuality at the cognitive level is conducted.

\vspace{-2mm}
\section{Methodology}
\label{sec:method}
Generally, an MLLM, denoted as $f_{\theta}(\cdot)$ and parameterized by $\theta$, takes the input question $Q$ and visual input $I$ as inputs, conducting reasoning across both modalities: $Y = f_{\theta}(I, Q)$, where $Y$ is the response.
To mitigate the hallucination in MLLMs, we perform a dedicated training-free reasoning framework that decomposes the one-step reasoning process into smaller sub-processes \cite{xu2024faithful,fei2024video}, adhering to the principle of progressing from perception to cognition. 
The overall structure of the proposed method is illustrated in Figure \ref{fig:framework}.
It consists of six reasoning modules: 
\textcircled{\small{\textbf{1}}} Target Identification and Visual Perception;
\textcircled{\small{\textbf{2}}} Visual Perception Verification;
\textcircled{\small{\textbf{3}}} Question Validation and Adjustment;
\textcircled{\small{\textbf{4}}} Commonsense Induction
\textcircled{\small{\textbf{5}}} Commonsense Verification;
\textcircled{\small{\textbf{6}}} Question answering.
In the following sections, we detail each reasoning module.

\paragraph{Target Identification and Visual Perception.}

In our first step, we expect to perceive the input text $T$ and vision $I$ by identifying key targets and focusing on visually relevant regions.
To accurately represent the perceived visual scene, we employ the SG, denoted $\mathcal{S}_g = \{\mathcal{O}, \mathcal{A}, \mathcal{R}\}$, a structured graph representation of visual scenes, which details not only objects $\mathcal{O}$ within the vision and their corresponding attributes $\mathcal{A}$ but also delineates the relationships $\mathcal{R}$ between objects.
Instead of relying on global perception, we prompt the model to infer only from the question what target objects are involved and what partial scene graph can be extracted from the image to answer the question.
Therefore, we construct the task prompt $P_{_{\scriptsize{\textcircled{\tiny{\bf{1}}}}}}$ (detailed in the Appendix \S D) combined with the input image $I$ to guide the MLLM to generate a partial SG ($\hat{\mathcal{S}}_g$) that is most related to answering the question:
\begin{equation}
\setlength{\abovedisplayskip}{1pt}
\setlength{\belowdisplayskip}{1pt}
    \hat{\mathcal{S}}_g = \{\hat{\mathcal{O}}, \hat{\mathcal{A}}, \hat{\mathcal{R}}\} = f_{\theta}(I, [P_{\scriptsize{\textcircled{\tiny{\bf{1}}}}};Q]),
\end{equation}
where $[;]$ denotes the concatenation operation
In practice, we transform the scene graph $\hat{\mathcal{S}}_g$ into JSON format, facilitating an easier interpretation and processing by the MLLMs.

\paragraph{Visual Perception Verification.}

In the second step, we verify the faithfulness of the partially extracted scene graph from the initial step to ensure accurate subsequent reasoning, concerning that MLLMs are susceptible to hallucinations.
We validate the three elements in the scene graph separately.
\textbf{Firstly}, we use the open-set object detection model, Grounding DINO \cite{abs-2303-05499}, to verify object fidelity; if the object fails to be detected, we classify the object as low-fidelity and remove it from the scene graph. 
\textbf{Secondly}, for attribute verification, we employ the Grounding DINO model, treating the attributes associated with objects as unified phrases for grounding. 
\textbf{Thirdly}, to ascertain the existence of relationships, we employ the similarity between the image union regions of the subject and object and relationship textual triplets using the BLIP \cite{0001LXH22}. 
By the above three steps, we can ensure the robustness and reliability of the SG for further analysis.
Through this module, the verified SG ($\bar{\mathcal{S}}_g = \{\bar{\mathcal{O}}, \bar{\mathcal{A}}, \bar{\mathcal{R}}\}$) will serve as low-level perception evidence (i.e., supporting rationale) for the next process of questioning verification and answering.

\vspace{-1mm}
\paragraph{Question Validation and Adjustment.}

In this process, we conduct a further examination of the input question to detect any inconsistencies with the high-fidelity visual perceptions established in the previous process. 
This verification is crucial, as indicated by our preliminary experiments in Appendix \S C.3, which reveal that approximately 47.8\% of response hallucinations are provoked by pre-existing hallucinations in the input questions, misleading the model into generating content that contradicts the visual facts.
For example, Case 3 in Figure \ref{fig:intro} demonstrates that the hallucination `cat' in the question misguided the erroneous generation of responses where the cat is invisible in the image.
To prevent such discrepancies from inducing further hallucinations, it is imperative to perform a de-confliction of the question.
Specifically, we categorize potential conflicts into three types based on the composition of the visual scene: \textbf{1)} \texttt{object conflict}, \textbf{2)} \texttt{attribute conflict}, and \textbf{3)} \texttt{relationship conflict}. 
We then employ in-context learning \cite{WeiTBRZBYBZMCHVLDF22} to prompt the model to review the input question regarding these three aspects. 
If conflicts are detected, the model is instructed to revise the question to rectify the inconsistency, striving to make minimal semantic alterations to maintain the question's integrity:
\begin{equation}
    \bar{Q} = f_{\theta}(I, [P_{\scriptsize{\textcircled{\tiny{\bf{3}}}}}; \bar{\mathcal{S}_g}; Q]),
\end{equation}
where $P_{_{\scriptsize{\textcircled{\tiny{\bf{3}}}}}}$ is the overall input prompt, and $\bar{Q}$ is the verified input question with a hint of whether the original questions contradict the visual content.
In addition, we discover that this module enables the model to engage in active thinking for questions that are unanswerable, ambiguous, or misleading, rather than simply responding with ``\emph{I don't know.}''

\vspace{-1mm}
\paragraph{Commonsense Induction.}

Having established a comprehensive understanding of the input query and its relevant visual perception, we can now consider answering the question. 
Although the model shows proficiency in handling perception-level queries based on the verified SG, it still exhibits hallucinations when responding to queries that require cognition-level reasoning, particularly those involving commonsense knowledge.
To mitigate this issue, we introduce a commonsense argumentation question-answering mechanism. 
Specifically, we harness the intrinsic self-analytic capabilities \cite{0002WSLCNCZ23} of the LLM to determine whether the question is answerable based on the available evidence. 
If it is unanswerable, the model outputs the necessary commonsense claims $\hat{\mathcal{C}} = \{\hat{c}_1, \cdot, \hat{c}_n\}$, which forms the fundamental basis required to answer the question:
\begin{equation}
    \hat{\mathcal{C}} = f_{\theta}(I, [P_{\scriptsize{\textcircled{\tiny{\bf{4}}}}}; \bar{\mathcal{S}}_g; \bar{Q}]),
\end{equation}
where $P_{\scriptsize{\textcircled{\tiny{\bf{4}}}}}$ is the task prompt. 
Through this module, we can obtain the {commonsense claims} $\hat{\mathcal{C}}$.

\vspace{-1mm}
\paragraph{Commonsense Verification.}
This process is designed to validate the faithfulness of the induced \texttt{[commonsense claims]}. 
Technically, we harness the Serper Google Search API to perform web searches using specific fact-based questions. 
By extracting and scrutinizing the top results, we retrieve a range of fact lists $\mathcal{R}$ from the API’s responses for analysis.
Then, we leverage the search results to verify the \texttt{[commonsense claims]} by prompting the model to categorize each claim $\hat{c}_i$ as either \textit{Hallucination} or \textit{Non-hallucination}:
\begin{equation}
\setlength{\abovedisplayskip}{1pt}
\setlength{\belowdisplayskip}{3pt}
    l_i = f_{\theta}([P_{\scriptsize{\textcircled{\tiny{\bf{5}}}}}; \hat{c}_i;\mathcal{R}]), 
\end{equation}
where $P_{\scriptsize{\textcircled{\tiny{\bf{5}}}}}$ is the overall input prompt for the subprocess-5, and $l_i$ is the label for the claim $\hat{c}_i$.
Then, we filter the hallucinated commonsense claims and finally obtain the verified commonsense claims $\bar{\mathcal{C}} = \{\bar{c}_1, \cdot, \bar{c}_m\}$.

\vspace{-1mm}
\paragraph{Question Answering.}
Finally, following the aforementioned processes, we have developed a comprehensive bottom-up understanding of the visual elements and questions. 
Thus, we now prompt the model to provide definitive answers to the question $\bar{Q}$ presented based on the given $I$ and its verified $\bar{S}_g$, and verified commonsense claims $\bar{C}$:
\begin{equation}
    Y = f_{\theta}(I, [P_{\scriptsize{\textcircled{\tiny{\bf{6}}}}};\bar{S}_g;\bar{C};\bar{Q}].
\end{equation}
Note that the task prompt $P_{\scriptsize{\textcircled{\tiny{\bf{6}}}}}$ 
slightly varies based on the information available.
With only perception-level content, we prompt MLLMs to determine whether it can answer the question. 
However, after verified commonsense is available, we assume sufficient accurate knowledge is obtained and prompt MLLMs to yield the final answer.

\begin{table*}[!t]
\centering
\fontsize{8}{9}\selectfont
\setlength{\tabcolsep}{1.7mm}
\begin{tabular}{lcccccccccccccccc|cc}
\toprule
\rowcolor[HTML]{FFFFFF} 
\multirow{2}{*}{Model} & \multicolumn{2}{c}{\textbf{OR}} & \multicolumn{2}{c}{\textbf{AR}} & \multicolumn{2}{c}{\textbf{SA}} & \multicolumn{2}{c}{\textbf{PR}} & \multicolumn{2}{c}{\textbf{C}} & \multicolumn{2}{c}{\textbf{PhD Avg.}} & \multicolumn{2}{c}{\textbf{WHOOPS!}} \\ 
\cmidrule(r){2-3} \cmidrule(r){4-5}\cmidrule(r){6-7}
\cmidrule(r){8-9} \cmidrule(r){10-11} \cmidrule(r){12-13} \cmidrule(r){14-15}
& \textbf{Neu.} & \textbf{Mis.} & \textbf{Neu.} & \textbf{Mis.} & \textbf{Neu.} & \textbf{Mis.} & \textbf{Neu.} & \textbf{Mis.} & \textbf{Neu.} & \textbf{Mis.} & \textbf{Neu.} & \textbf{Mis.} & \textbf{VQA} & \textbf{Gen.}  \\ 
\midrule
\textbf{LLaVA-1.5} & 65.9 & 22.5 & 62.6 & 11.8 & 69.0 & 32.8 & 47.9 & 14.5 & 47.3 & 11.7   & 58.5 & 18.7 & 47.3  & 67.9 \\
\rowcolor{lightblue} \quad {+ Ours} & 67.5 & 35.4 & 67.0 & 24.3 & 76.3 & 46.5 & 53.3 & 29.0 & 52.1 & 19.8    & 63.2 (\textcolor{red}{+4.7}) & 31.0 (\textcolor{red}{+12.3}) & 54.5 (\textcolor{red}{+7.2})  & 72.3 (\textcolor{red}{+4.4}) \\
 \midrule
\textbf{Qwen-VL-Chat} & 79.5 & 46.3 & 80.9 & 42.1 & 73.6 & 37.9 & 69.1 & 43.1 & 57.6 & 32.8   & 72.1 & 40.4 & 48.7  & 67.5 \\
 \rowcolor{lightblue} \quad {+ Ours} & 81.8 & 56.8 & 86.9 & 54.0 & 77.4 & 47.3 & 83.4 & 62.4 & 64.1 & 42.3   & 78.7 (\textcolor{red}{+6.6}) & 52.5 (\textcolor{red}{+12.1}) & 54.3  (\textcolor{red}{+5.6})  &  73.4   (\textcolor{red}{+5.9}) \\
  \midrule
 \bf MiniGPT-V2 & 84.5 & 43.3 & 71.5 & 26.1 & 78.1 & 20.5 & 62.7 & 35.3 & 66.1 & 28.7   & 72.6 & 30.8 & 49.1 &  71.3 \\
 \rowcolor{lightblue} \quad {+ Ours} & 86.0 & 59.9 & 76.1 & 38.0 & 79.9 & 42.4 & 71.0 & 61.8 & 68.2 & 48.8    & 76.2 (\textcolor{red}{+3.6}) &  50.2 (\textcolor{red}{+19.4}) & 51.6 (\textcolor{red}{+2.5})  & 75.6  (\textcolor{red}{+4.3}) \\
  \midrule
\textbf{GPT-4V} & 83.2 & 76.4 & 76.2 & 28.6 & 76.0 & 47.2 & 59.7 & 42.5 & 57.6 & 40.6  & 70.5 & 47.1 & 64.8 & 81.7 \\
 \rowcolor{lightblue} \quad {+ Ours} & 88.0 & 87.3 & 85.8 & 43.7 & 81.7 & 65.5 & 61.8 & 52.0 & 85.0 & 75.4    & 80.5(\textcolor{red}{+10.0}) & 64.8 (\textcolor{red}{+17.7}) &  69.9 (\textcolor{red}{+5.1})  & 89.8  (\textcolor{red}{+8.1}) \\
\bottomrule
\end{tabular}
\vspace{-2mm}
\caption{Evaluation on PhD and WHOOPS! benchmarks. The PhD dataset is split into neural (\textbf{Neu.}), and misleading (\textbf{Mis.}) questions in Object Recognition (\textbf{OR}), Attribute Recognition (\textbf{AR}), Sentiment Analysis (\textbf{SA}), and Positional Reasoning (\textbf{PR}), and Counting (\textbf{C}). \textbf{PhD Avg.} denotes the average performance on the PhD dataset. For the WHOOPS! benchmark, we evaluate our method on the compositional \textbf{VQA} and explanation generation (\textbf{Gen.}) tasks.}
\label{tab:phd}
\vspace{-2mm}
\end{table*}

\begin{table}[!th]
\centering
\fontsize{8.5}{9}\selectfont
\setlength{\tabcolsep}{1.0mm}
\begin{tabular}{@{}lccccc@{}}
\toprule
{Model}  & \textbf{Acc.} & \textbf{Prec.} & \textbf{Rec. }& \textbf{F1} & \textbf{Yes}  \\ 
\midrule
\textbf{LLaVA-1.5}$^*$ & 84.1 & 90.9 & 75.7 & 82.6 & 41.7  \\ 
\quad + Woodpecker$^\flat$ & 85.7 &  91.6  & 76.3  & 83.2 & - \\
\quad + LURE$^\dag$  & 84.5 &  & - & 85.0 & - \\
\quad + LogicCheckGPT$^\dag$ & 90.0 & - & - & 89.0 & - \\
\quad + DVP$^*$ & 85.7 & 95.5 & 74.9 & 84.0 & 38.4  \\
\rowcolor{lightblue} \quad + Ours & {91.2} & {96.1} & {87.4} & {91.5}(\textcolor{red}{+2.5}) & 43.5   \\
\midrule
\textbf{Qwen-VL-Chat}$^*$ & 84.3 & 94.2 & 73.0 & 82.3 & 38.7  \\
\quad + Woodpecker$^\flat$ & 85.7 &  93.6  & 76.3  & 84.0 & - \\
\quad + LURE$^\flat$  & 86.7 & 93.4 & 79.0 & 85.5 & - \\
\quad + LogicCheckGPT$^\flat$ & 87.6 & 91.0 & 83.9 & 87.3 & - \\
\quad + DVP$^*$ & 86.3 & 99.6 & 72.8 & 84.1 & 36.5 \\
\rowcolor{lightblue} \quad + Ours & {89.7} & {98.6} & {86.5} & {92.2}(\textcolor{red}{+4.9}) & 41.2 \\
\midrule
\textbf{GPT-4V}$^*$ & 82.7 & 85.5 & 78.8 & 82.0 & 46.1  \\
\quad + Woodpecker$^\flat$ & 83.1 &  86.2  & 79.3  & 82.6 & - \\
\quad + LURE$^\flat$  & 84.4 & 86.9 & 81.6 & 84.2 & - \\
\quad + LogicCheckGPT$^\flat$ & 85.9 & 87.3 & 83.1 & 85.1 & - \\
\quad + DVP$^*$ & 86.8 & 88.2 & 85.0 & 86.6 & 48.2  \\

\rowcolor{lightblue} \quad + Ours & {94.2} & {98.8} & {89.5} & {93.9}(\textcolor{red}{+7.3}) & 47.7 \\
\bottomrule
\end{tabular}
\vspace{-2mm}
\caption{Performance Evaluation of models on POPE (Adversarial) Setting. the scores with * are derived from \cite{abs-2403-13513}, $^\dag$ are copied from \cite{abs-2402-11622}, $^\flat$ are re-implemented based on the open-source code. }
\label{tab:pope}
\vspace{-4mm}
\end{table}

\vspace{-2mm}
\section{Settings}
\label{settings}

\vspace{-1mm}
\paragraph{Datasets and Baselines.}

To rigorously evaluate the performance of the proposed framework, we selected two categories of benchmarks based on the levels at which hallucinations typically occur:
\texttt{1) Perception-level benchmarks} are used to test the model's ability to de-hallucinate visual content concerning objects, attributes, and relationships. This includes benchmarks such as POPE \cite{LiDZWZW23}, PHD \cite{abs-2403-11116}, AMBER \cite{abs-2311-07397} and WHOOPS!-VQA \cite{GuettaBHSES023}.
\texttt{2) Cognition-level benchmarks }are aimed at evaluating the model on more complex issues, such as unanswerable or ambiguous questions, or those requiring commonsense knowledge, such as sentiment analysis. 
For this purpose, we selected representative datasets like WHOOPS!-Gen \cite{GuettaBHSES023} and VQAv2-IDK \cite{abs-2402-09717}.
To ensure a comprehensive and impactful assessment, we chose a lineup of representative and widely recognized models, including five Multimodal Large Language Models (MLLMs): LLaVA-1.5 \cite{abs-2310-03744}, MiniGPT-v2 \cite{abs-2310-09478}, Qwen-VL\cite{abs-2308-12966}, and GPT-4V \cite{gpt4V} (model: gpt-4-vision-preview).
We also compare our method with advanced hallucination detection and mitigation methods, including Woodpecker \cite{abs-2310-16045}, LURE \cite{abs-2310-00754}, LogicCheckGPT \cite{abs-2402-11622}, and DVP \cite{abs-2403-13513}.

\vspace{-1mm}
\paragraph{Implementation and Evaluation Metrics.}
Our framework operates without training, leveraging an open-source pre-trained model to assess performance. 
We employ Grounding DINO \cite{abs-2303-05499} for object and attribute verification and BLIP \cite{0001LXH22} for validating the existence of relationships. 
To quantify the evaluation, we use accuracy as our metric on the PhD datasets, as followed in \cite{abs-2403-11116}.
For the POPE dataset, we utilize a range of evaluation metrics including Accuracy (Acc.), Precision (Pre.), Recall (Rec.), F1 score (F1), and Yes. 
Following \cite{abs-2402-09717}, we calculate the IDK Metric to assess performance on the VQAV2-IDK dataset.
Additionally, the AMBER \cite{abs-2311-07397} score and F1 score are computed to evaluate outcomes on the AMBER and Hal-Eval datasets.

\begin{table}
\centering
    \fontsize{8}{10}\selectfont
    \setlength{\tabcolsep}{1.3mm}
    \begin{tabular}{@{}lccccc@{}}
    \toprule
    \multirow{1}{*}{Model} & \multicolumn{1}{c}{\textbf{Uans}} & \multicolumn{1}{c}{\textbf{FQ}} & \multicolumn{1}{c}{\textbf{DK}} & \multicolumn{1}{c}{\textbf{NS}} & \multicolumn{1}{c}{\textbf{Total}}  \\  
    \midrule
    \bf LLaVA-1.5$^*$ & 4.71 & 1.62 & 5.66 & 0.41 &  3.73 \\
     \rowcolor{lightblue}\quad + ours & 24.67 & 8.66 & 10.83 & 17.69 & 15.08 (\textcolor{red}{+11.35})  \\ 
     \midrule
    \bf MiniGPT-V2 & 5.13  & 1.57 & 6.70 & 0.82  & 3.86\\ 
     \rowcolor{lightblue}\quad + ours & 25.02 & 10.62 & 18.22 & 19.21 & 18.97(\textcolor{red}{+15.11})  \\ 
     \midrule
    \bf GPT-4V$^*$ & 52.02 & 30.62 & 49.22 & 42.21 & 41.97 \\
     \rowcolor{lightblue} \quad + ours & 60.53& 33.01 & 62.85 & 67.18 & 55.99(\textcolor{red}{+14.02}) \\ 
    \bottomrule
    \end{tabular}
    \vspace{-2mm}
    \caption{Evaluation on VQAv2-IDK dataset. The score with * are copied from \cite{abs-2402-09717}, `Uans', `FQ', `DK', `NS' denotes `Unanswerable', `False Questions', `Don’t Know', `Not Sure', respectively.}
    \label{tab:vqav2-idk}
    \vspace{-2mm}
\end{table}

\begin{table}[]
    \centering
    \fontsize{8}{10}\selectfont
    \setlength{\tabcolsep}{1.8mm}
    \begin{tabular}{@{}lcccccc@{}}
    \toprule
    \multirow{1}{*}{Model} & \multicolumn{1}{c}{\textbf{OR}} & \multicolumn{1}{c}{\textbf{AR}} & \multicolumn{1}{c}{\textbf{PR}} & \multicolumn{1}{c}{\textbf{SA}} & \multicolumn{1}{c}{\textbf{ VQA}} &  \multicolumn{1}{c}{\textbf{Gen.}} \\  
    \midrule
    \rowcolor{lightblue}  \bf LLaVA-1.5(Ours) & 35.4 & 24.3 & 46.5 & 29.0 & 64.5 & 74.3  \\
    \midrule
    \quad w/o TIVP\&VPV & 26.6 & 15.3 & 33.2 & 6.8 & 50.6 & 68.7 \\
    \quad w/o VPV  & 27.6 & 17.3 & 34.3 & 12.3  & 58.7 & 69.1 \\
    \quad w/o QAV  & 24.1 & 14.1  & 38.2 &  15.4 & 63.0 & 71.8 \\ 
    \cdashline{1-7}
    \quad w/o CV  & 31.1 & 20.4 & 45.0 & 25.2 & 62.8 & 70.9 \\
    \quad w/o CV\&CI  & 30.9 & 21.8 & 42.5 & 18.7 & 60.3 & 71.0 \\ 
    \bottomrule
    \end{tabular}
    \vspace{-2mm}
    \caption{Ablation study on PhD (misleading questions) and WHOOPS! dataset to validate the efficacy of each module in mitigating hallucination at perception and cognition levels.}
    \label{tab:ablation}
    \vspace{-3mm}
\end{table}

\section{Experiments and Analysis}
\label{exp_analysis}

\paragraph{Results on Perception Hallucination.}

We first evaluate our model's performance in mitigating perception-level hallucinations using PhD and POPE datasets.
The experimental results, as shown in Tables \ref{tab:phd} and \ref{tab:pope} reveal a notable performance improvement in our model compared to baseline MLLMs. 
Specifically, our model achieves an average improvement of 10.8\% on the PhD, 4.9\% on the POPE, and 5.1\% on the WHOOPS!-VQA, with GPT-4V exhibiting a marked enhancement following the integration of our proposed framework.
Our model's performance is particularly pronounced when addressing misleading questions, where input questions contain conflicts, compared to neural questions.
Moreover, when juxtaposed with baselines designed to mitigate hallucinations, our method consistently displays distinct advantages. 
Notably, in contrast to the training-dependent LURE model, our approach operates on a training-free basis and still achieves superior performance.
Additionally, our method, designed holistically, outperforms traditional post-hoc correction approaches, such as Woodpecker, which typically verify only outputs, leading to notably better performance outcomes.

\paragraph{Results on Cognition Hallucination.}

Here, we validate the model's ability to mitigate hallucinations at the cognition level. 
We perform experiments on VQAv2-IDK and WHOOPS!-Gen benchmarks, where the questions require not only perceptual abilities but also cognition-level reasoning skills. 
As shown in Table \ref{tab:phd} and \ref{tab:vqav2-idk}, integrating our proposed mechanism leads to a significant performance improvement across various MLLMs, achieving an average of 13.49\% improvement on VQAv2-IDK and 6.18\% on WHOOPS!-Gen, showcasing the effectiveness of our approach in reducing cognition-level hallucinations.

\paragraph{Ablation Study.}

To directly assess the contribution of each module in our framework, we conduct an ablation study. 
The results are detailed in Table \ref{tab:ablation}. 
Firstly, the removal of any module results in a decline in model performance. 
Most notably, omitting the visual perception (\texttt{TIVP}) and verification (\texttt{VPV}) modules lead to the most significant performance deterioration, regardless of whether the questions are at the perception or cognition level. 
This underscores the importance of accurate perception and verification of image content in vision-language understanding.
Furthermore, while eliminating the commonsense induction (\texttt{CI}) and verification (\texttt{CV}) modules do not significantly impact performance on datasets that primarily focus on perceptual abilities,  it significantly affects tasks requiring higher-level reasoning.
Specifically, for tasks such as sentiment analysis (\textbf{}{SA}) or commonsense violation explanation (\textbf{Gen.}), which cannot be inferred from visual perception alone, the absence of these modules leads to notable performance degradation.

\begin{figure}
    \centering
    \includegraphics[width=0.78\columnwidth]{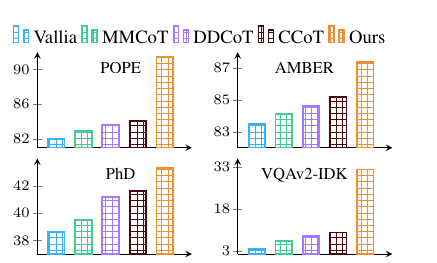}
    \vspace{-2mm}
    \caption{The comparison of different CoT mechanisms.}
    \label{fig:cot}
    \vspace{-2mm}
\end{figure}

\paragraph{The Impact of Various CoT.}

We examine the advantages of our CoT-based framework compared to other CoT frameworks. 
To this end, we compared our framework against three other CoT-baed methods: MMCoT \cite{abs-2302-00923}, DDCoT \cite{ZhengYTZY23}, and CCoT \cite{abs-2311-17076}. 
As illustrated in Figure \ref{fig:cot}, it is evident that the design of a CoT framework can significantly reduce model hallucinations.
Moreover, while the CCoT framework straightforwardly re-enters an induced scene graph for reasoning, our model incorporates additional subprocesses such as perception verification and input question validation, leading to a significant enhancement in mitigating hallucinations.

\begin{table}[!t]
    \centering
    \fontsize{8}{9}\selectfont
    \setlength{\tabcolsep}{1.5mm}
    \caption{Comparison of two MLLMs and one specialist on SG generation task.}
    \label{tab:sgg}
    \vspace{-2mm}
    \begin{tabular}{@{}lccc@{}}
    \toprule
    \multirow{1}{*}{Model} & \multicolumn{1}{c}{\textbf{Object}} & \multicolumn{1}{c}{\textbf{Attribute}} & \multicolumn{1}{c}{\textbf{Relation}}   \\  
    \midrule
    LLaVA-1.5 & 75.6 & 72.6 & 62.3   \\
    GPT-4 & 86.7 & 83.1 & 76.0   \\ 
    \midrule
    HiKER-SGG \cite{abs-2403-12033} & 74.6 & -  & 67.9  \\ 
    \bottomrule
    \end{tabular}
    \vspace{-2mm}
\end{table}

\begin{figure}[!t]
\begin{center}
    \includegraphics[width=0.5\columnwidth]{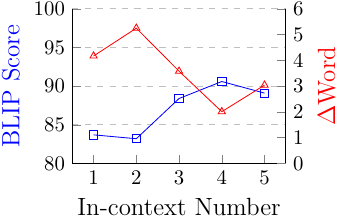}
\end{center}
\vspace{-4mm}
\caption{Conflict resolution performance with varying numbers of in-context examples.}
\label{fig:input-modi}
\vspace{-5mm}
\end{figure}

\vspace{-1mm}
\paragraph{Scene Graph Creation and Verification Outcomes.}

Next, we comprehensively evaluate the impact of SG generation and verification within our framework.
Initially, we compare the performance of the MLLMs with an existing SoTA specialist HiKER-SGG \cite{abs-2403-12033} in SG extraction. 
Specifically, we randomly select 1k images from the Visual Genome \citep[VG;][]{KrishnaZGJHKCKL17} benchmark, prompt MLLMs to extract a complete scene graph for each image, comparing the results with those from HiKER-SGG. 
Given that MLLMs operate in an open-vocabulary setting, while the baseline is confined to a closed set, we conduct a mapping of relationships to calculate the final outcomes.
As depicted in Table \ref{tab:sgg}, MLLMs outperform the HiKER-SGG, particularly as the specialist models are limited to SG generation within a closed vocabulary.

\begin{figure*}[!t]
\centering
\includegraphics[width=1.0\textwidth]{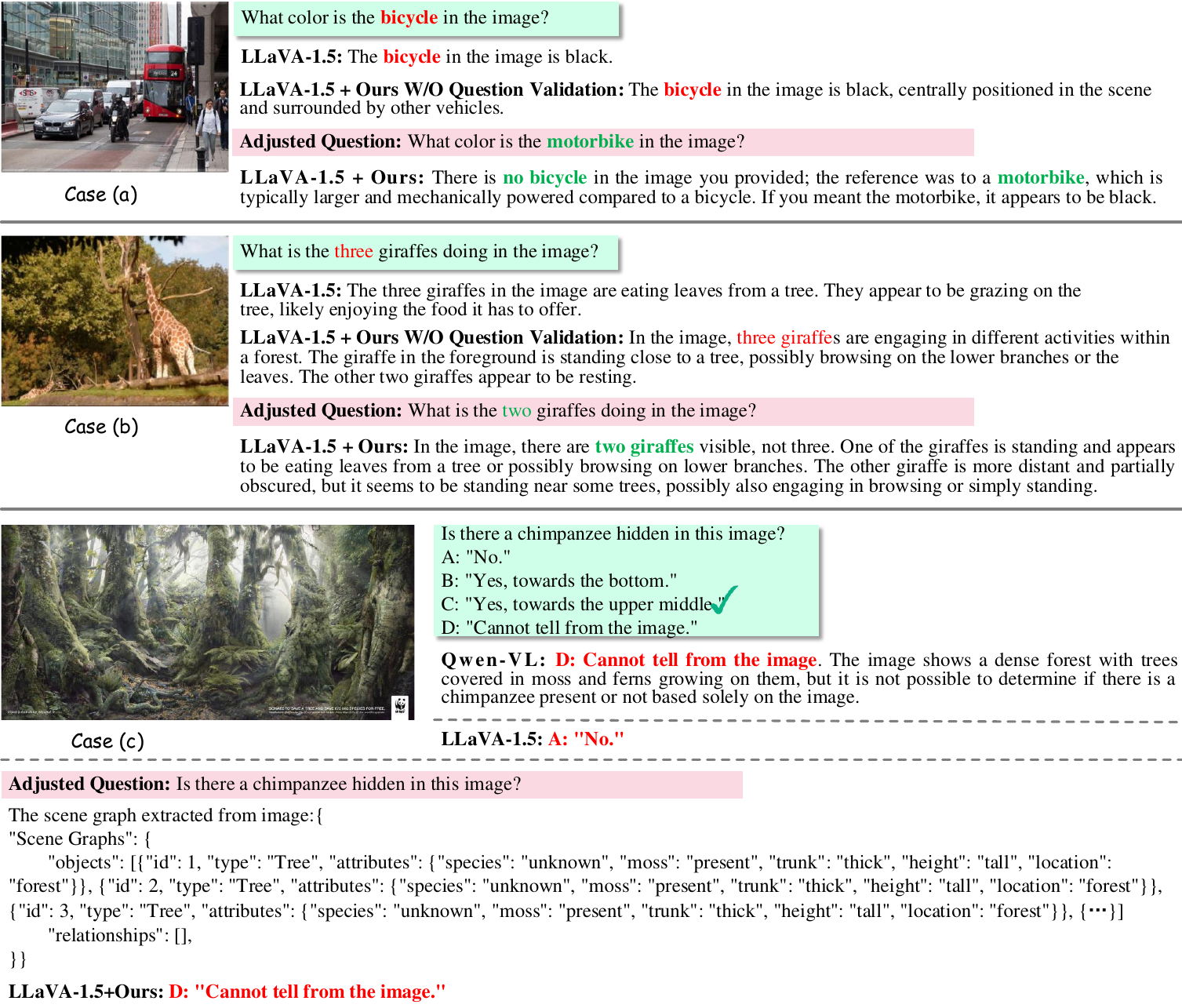}
\vspace{-5mm}
\caption{
Illustration of example outputs. Case (a) and (b)  outputs with and without question validation for input questions containing conflicts. Hallucinations are highlighted in \textcolor{red}{\textbf{red}} and non-hallucinated in \textcolor{defgreenDark}{\textbf{green}}. The input raw questions are marked in \colorbox{defgreenLight}{green}, and the adjusted questions in \colorbox{defredLight}{red}.
Case (c) shows a failure example. For more results, refer to the Appendix \S F.2.
}
\vspace{-5mm}
\label{fig:input-conf}
\end{figure*}

\vspace{-1mm}
\paragraph{Effectiveness of Conflict Resolution in Input question.}

Next, we examine whether resolving conflicts in input questions enhances reasoning and reduces hallucinations. We conduct experiments on the AMBER dataset, evaluating performance with varying numbers of in-context examples and measuring both the BLIP score \cite{0001LXH22} and the average number of word modifications before and after sentence editing.
As shown in Figure \ref{fig:input-modi}, our method modifies an average of two words in each input question before and after editing, while maintaining high semantic similarity, indicating that our interventions can de-conflict model inputs without drastically changing the implicit intent comprehension. 
Furthermore, by visualizing sentences and their corresponding responses both before and after editing in Figure \ref{fig:input-conf}, we observe that post-adjusted inputs align more closely with the image, such as minor changes in quantities and objects in images.
Most importantly, these modifications ensure that the model's responses are consistent with the visual information provided, effectively mitigating hallucinations.

\vspace{-1mm}
\paragraph{Case Study.}
We provide examples to intuitively demonstrate the effectiveness of our proposed framework, as illustrated in Figure \ref{fig:input-conf}. 
Incorporating the proposed framework significantly enhances the proficiency of our model in mitigating hallucinations in input queries and enables it to respond with more detailed rationales. 
However, as shown in Figure \ref{fig:input-conf} (c), we observe that some limitations remain, particularly in cases involving ambiguous images where the model's visual information processing capabilities are insufficient, leading to failure in some responses.

\vspace{-2mm}
\section{Conclusion}

In this work, we introduce a holistic, training-free reasoning framework to mitigate hallucinations in MLLMs. 
This framework emulates the human problem-solving process by dividing reasoning into six sub-processes, from perception-level visual understanding to cognition-level commonsense reasoning. 
Technically, we design a synergistic approach that incorporates perception and cognition-level understanding alongside verification.
Additionally, we innovatively propose a question reconsideration and rectification mechanism. 
Extensive experiments across six benchmarks show that integrating our method into various MLLMs consistently enhances performance on perception and cognition-level questions. 
Furthermore, in-depth analyses and visualizations reveal that the framework effectively identifies and reduces conflicts between input visual content and questions.
The incorporation of verified commonsense further remarkably reduces hallucinations in responses.

\bibliography{aaai25}

\newpage
\appendix

\section{Ethic Statement}
\label{ethic_statement}

\paragraph{Transparency and Integrity.}
We ensure that all methodologies, data sources, and technologies used in this study are disclosed transparently. We aim to provide a comprehensive and honest account of our findings, acknowledging both the capabilities and limitations of our proposed solution.

\paragraph{Data Privacy and Security.} 
Our research utilizes datasets that are either publicly available or collected with explicit consent. We adhere to strict data privacy and security protocols to protect the information and ensure it is used solely for the purposes of this research.

\paragraph{Bias Mitigation.}
Recognizing the potential for bias in AI models, particularly in vision-language tasks, we rigorously test our framework across diverse datasets. 
This approach is designed to identify and mitigate biases that may affect the model’s performance or lead to unfair outcomes in its applications.

\section{Limitation}
\label{limitation}

Despite the promising results, our proposed bottom-up holistic reasoning framework has certain limitations that warrant further investigation:

\paragraph{Computational Complexity.} The multi-step reasoning process introduced in this framework may affect the model's efficacy, especially when handling simple questions in quick-response scenarios.

\paragraph{Self-knowledge Capacity.}
Our model may be limited by the inherent understanding capabilities of existing MLLMs, especially when dealing with images that feature ambiguity and abstract scenes. 
However, our model strives to provide answers by employing a verification mechanism that significantly reduces hallucinations.

\paragraph{Evaluation Metrics.}
Current evaluation metrics for hallucination might not fully capture the nuanced improvements provided by our framework. Developing more comprehensive and robust metrics is essential for accurately assessing the hallucination rate.

\section{Preliminary}

Here, we provide several important preliminary contents as a supplement to the main article.

\subsection{Hallucination Types}
We summarize the types of hallucination in the following types:

\begin{itemize}
    \item \textbf{Perception-level Hallucination}: focuses on hallucinations that are primarily sensory and are linked directly to how objects, attributes, and relationships are perceived.
    \begin{itemize}
        \item \textbf{Object Hallucination}: The erroneous perception of non-existent object categories or incorrect categories of MLLMs that are not present in the given image. 

        \item \textbf{Attribute Hallucination}: 
        The misattributing characteristics to objects, such as color, shape, material, content, counting, action, etc.

        \item \textbf{Relation Hallucination}: The incorrectly perceiving relationships between objects (such as human-object interactions or relative positions) do not align with the actual image content.

        \item \textbf{Object Confliction}: refer to the input question containing the objects that are not presented in the given image.  

        \item \textbf{Attribute Conflication}: refer to the input question containing the attributes that are not presented in the given image.

        \item \textbf{Relation Conflication}:  refer to the input question containing the relationships that are not presented in the given image.

         \end{itemize}
    \item  \textbf{Cognition-level Hallucination}: focuses on hallucinations that are more about errors in reasoning, memory, and the processing of knowledge, rather than direct sensory errors.
    \begin{itemize}
        \item \textbf{Commnonsense Hallucination}: involves a failure to apply everyday knowledge and logical reasoning to assess situations or interpret phenomena correctly.
        \item \textbf{Knowledge Overgeneralization}:  refers to the inappropriate application of specific knowledge to broader contexts, which is incorrect and leads to inaccurate assumptions or conclusions.
    \end{itemize}
\end{itemize}

\subsection{Scene Graph Representation}

A scene graph (SG) is a structured representation of the visual scene within an image.
Specifically, it consists of the following components:

\begin{itemize}
    \item \textbf{Nodes}: Each node in a scene graph represents an object present in the scene. Nodes are often labeled with categories such as `person', `car', `tree', etc. These categories help in identifying what the objects are.

    \item \textbf{Attributes}: Nodes may have attributes that describe their properties, such as color, size, state, or any other relevant characteristics. For instance, a `car' node might have attributes like `red' and `convertible'.

    \item \textbf{Edges}: Edges connect the nodes and represent the relationships between them. These relationships can be spatial (e.g., next to, above, behind) or interactive (e.g., riding, holding). This helps in understanding how objects are situated or interact with each other within the scene.
\end{itemize}

Recently, SGs have been introduced as powerful vision-language representations, and have been extensively explored in many existing works \cite{Wu0ZC23,QianCCWJ23,DhamoFLNHT020}.

\begin{table}[ht]
\centering
\fontsize{8.0}{10}\selectfont
\setlength{\tabcolsep}{0.7mm}
\caption{Comparative statistics of hallucination (Hallu.) and non-hallucination (Non-Hallu.) instances across three benchmark datasets and performance metrics for various MLLMs.}
\label{tab:input-hallu-static}
\begin{tabular}{@{}lcccccc@{}}
\toprule
& \multicolumn{2}{c}{PhD} & \multicolumn{2}{c}{VQAv2-IDK (val)} & \multicolumn{2}{c}{AMBER} \\
\cmidrule(lr){2-3} \cmidrule(lr){4-5} \cmidrule(lr){6-7}
 & Hallu & Non-Hallu. & Hallu. & Non-Hallu. & Hallu. & Non-Hallu. \\
\midrule
\textbf{Proportion(\%)} & 70.15 & 29.84 & 31.75 & 68.25 & 41.50 & 58.50 \\
\bottomrule
\toprule
Model & \multicolumn{2}{c}{F1 $\uparrow$} & \multicolumn{2}{c}{IDK score$\uparrow$}  &  \multicolumn{2}{c}{AMBER$\uparrow$}  \\
\cmidrule(r){1-1} \cmidrule(r){2-3} \cmidrule(r){4-5} \cmidrule(r){6-7} 
LLaVA-1.5 & 18.66 & 58.54 & 1.62 & 3.59 & 6.15 & 8.23 \\
Qwen-VL-Chat & 40.44 & 72.14 & 1.57 & 4.56 & 2.56 & 6.37 \\
 MiniGPT-V2& 30.78 & 72.58 & 2.34 & 6.78 & 6.45 & 19.20 \\
 GPT-4V & 47.06 & 70.54 & 30.62 & 47.81 & 3.11 & 5.23 \\
\bottomrule
\end{tabular}
\end{table}

\subsection{The Impact of the Input Hallucination}
\label{append_input_hallu}

Here, we examine the impact of input hallucinations on response hallucinations. 
Our initial analysis of existing benchmarks reveals that 47.8\% of the data exhibit this phenomenon, as shown in Table \ref{tab:input-hallu-static}.
Moreover, it is a significant prevalence in the common real-world scenario where users inadvertently provide erroneous queries to models.
Subsequently, we sample an equivalent number of instances from both hallucination and non-hallucination datasets to evaluate the performance of various MLLMs. 
As it can be seen from Table \ref{tab:input-hallu-static}, all MLLMs across three datasets demonstrate a marked performance degradation when faced with inputs containing hallucinations.
This underscores the critical importance of recognizing input hallucinations in addressing the issue of response hallucinations, confirming that perception of input flaws is vital for mitigating erroneous model responses.

\section{Detailed Input Prompt}
\label{detailed_input_prompt}

Here, we provide detailed task prompts, as well as their inputs and outputs, for each module of the de-hallucination reasoning framework.
The task prompts in each MLLM may differ slightly. 
\newpage

\begin{instructionbox}{$P_{\textcircled{\small{\textbf{1}}}}$ Target Identification and Visual Perception}
{}[\texttt{System Prompt}] \\
You are an advanced and intelligent expert equipped with proficiency in analyzing image data. You can provide helpful and detailed answers to the user's questions.

[\texttt{Instruction}]\\
Given the image and its associated question, you need to identify the target objects involved in the question and extract a partial scene graph in JSON format that is most related to answering the question based on the image.
This graph should include the following elements:\\
1. Objects that are crucial for answering the question.\\
2. Relevant attributes of these objects.\\
3. Relationships between these objects that are significant for the question.\\
The structure of the JSON scene graph should be the following format:\\
    \{\\
"Scene Graphs": \\
"objects": [\\
    \{
    "id": 1,
    "type": "Hat",
    "attributes": \{"color": blue, "text": "LOVE"\}
    \},\\
      ...\\
    ],\\
    "relationships": [\\
      \{
        "source": 1,
        "target": 2,
        "relation": "above"
      \}\\
      ...\\
    ]\\
\}\\
\hdashrule[0.5ex]{\linewidth}{0.2pt}{2mm 1mm} 
Given \image and [\texttt{Question}], please provide your detailed analysis. 
\end{instructionbox}

\begin{instructionbox}{$P_{\textcircled{\small{\textbf{3}}}}$ Question Validation and Adjustment}
{}[\texttt{System Prompt}] \\
You are an advanced and intelligent expert equipped with proficiency in analyzing image data. You can provide helpful and detailed answers to the user's questions.

[\texttt{Instruction}]\\
Given the image and its associated scene graph, please analyze if there is any potential conflict between the visual content and question in the following three aspects:\\
\quad 1. Object Conflict: The question might mention an object that does not exist in the image.\\
\quad 2. Object Attribute Conflict: The attributes of an object described in the question may not align with those depicted in the image.\\
\quad 3. Object Relation Conflict: The relationships between objects described in the question may differ from those observed in the image.\\
Next, I will provide several examples to help you understand this task more clearly:\\
\hdashrule[0.5ex]{\linewidth}{0.2pt}{2mm 1mm} 
\textit{\textbf{[Example-1]}}\\
{}[\texttt{input}] 
\image \\
{}[\emph{Scene Graph}]
\{
  "Scene Graphs": \{\\
    "objects": [\\
      \{
        "id": 1, "type": "Person", "attributes": \{"activity": "holding surfboard", "attire": "wetsuit"\}
     \},\\
      \{
        "id": 2, "type": "Person", "attributes": \{"activity": "standing", "attire": "blue t-shirt", "accessory": "backpack"\}
      \},\\
      \{
        "id": 3, "type": "Person", "attributes": \{"activity": "standing", "attire": "green t-shirt", "accessory": "backpack"\}
      \},\\
      \{
        "id": 4, "type": "Person", "attributes": \{"activity": "holding paddle", "attire": "shorts"\}
      \}\\
    ],\\
    "relationships": [\\
      \{
        "source": 1, "target": 4, "relation": "next to"
      \},\\
      \{
        "source": 2, "target": 3, "relation": "next to"
      \}\\
    ]
  \}
\}\\
{}[\textit{Question}]: "Are there four persons on the beach?" \\
{}[\textit{Output}]
\\
Analyzing the potential conflicts between the scene graph derived from the image and the associated question "Are there four persons on the beach?" involves examining the three specified aspects:\\
1. Object Conflict: The question references "four persons on the beach." ... \\
2. Object Attribute Conflict: The attributes given for each ... \\
3. Object Relation Conflict: The relationships noted in the scene graph ... \\
\textbf{Conclusion}: There is no conflict between the visual content in the image and the question in terms of the object count, their attributes, or their relationships as analyzed from the scene graph.\\
\textbf{Output Question}: "Are there four persons on the beach?"\\
\hdashrule[0.5ex]{\linewidth}{0.2pt}{2mm 1mm} 
\textit{\textbf{[Example-2]}}\\
...\\
\hdashrule[0.5ex]{\linewidth}{0.2pt}{2mm 1mm} 
{}[\texttt{Input}]\\
Given the \image and associated \SG, please verify the [\texttt{Question}] and provide a detailed analysis.
\end{instructionbox}

\begin{instructionbox}{$P_{\textcircled{\small{\textbf{6}}}}$  Question Answering}
{}[\texttt{System Prompt}] \\
You are an expert with an extensive amount of common sense knowledge and are equipped with proficiency in analyzing image data. You can provide helpful and detailed answers to the user's questions.

[\texttt{Instruction}]\\
Given the image and its associated scene graph, you need to first determine whether the content available through visual perception is sufficient to answer the given question. If it is possible, you need to provide an answer. 
\hdashrule[0.5ex]{\linewidth}{0.2pt}{2mm 1mm} 
If not, output the question is unanswerable. \\
Given \image and \SG, please give your analysis of the 
[\texttt{Question}]:
\end{instructionbox}

\begin{instructionbox}{$P_{\textcircled{\small{\textbf{4}}}}$ Commonsense Induction}
{}[\texttt{System Prompt}] \\
You are an expert with an extensive amount of common sense knowledge and are equipped with proficiency in analyzing image data. You are able to give helpful and detailed answers to the user's questions.

[\texttt{Instruction}]\\
Given the image and its associated scene graph, you need to deduce the commonsense claim required to answer the question in the following format:\\
{}[\texttt{Commonsense Claims}] \\
{}[\texttt{Claim 1}]: $\cdots$ \\
{}[\texttt{Claim 2}]: $\cdots$ \\
$\cdots$
\\
\hdashrule[0.5ex]{\linewidth}{0.2pt}{2mm 1mm} 
Given \image and \SG, please give your analysis of the 
[\texttt{Question}]:
\end{instructionbox}

\begin{instructionbox}{$P_{\textcircled{\small{\textbf{5}}}}$ Commonsense Verification}
{}[\texttt{System Prompt}] \\
You are an advanced and intelligent expert equipped with proficiency in analyzing image data. You can provide helpful, and detailed answers to the user's questions.

[\texttt{Instruction}]\\
Given the commonsense claim list and their corresponding retrieved factual lists, you need to verify the faithfulness of each claim and clarify them into  \textit{Hallucination} or \textit{Non-hallucination}.
The final answer should be in the following format:\\
{}[\texttt{Claim 1}]: $\cdots$\\
{}[\texttt{Label}]: Non-hallucination \\
{}[\texttt{Claim 1}]: $\cdots$\\
{}[\texttt{Label}]: Hallucination \\
$\cdots$\\
\hdashrule[0.5ex]{\linewidth}{0.2pt}{2mm 1mm}
Give the [\texttt{commonsense claim}] list and \texttt{retrieval factual lists}], provide your analysis for each claim:

\end{instructionbox}

\begin{instructionbox}{$P_{\textcircled{\small{\textbf{6}}}}$  Question Answering}
{}[\texttt{System Prompt}] \\
You are an advanced and intelligent expert equipped with proficiency in analyzing image data. You are able to give helpful and detailed answers to the user's questions.

[\texttt{Instruction}]\\
Given the \image and its associated \SG, and verified [\texttt{verified commonsense claims}], please answer the [\texttt{Question}]. 
\end{instructionbox}

\section{Extended Experiment Configurations}
\label{extended_exp_config}

\subsection{Detailed Datasets}

Here, we give a detailed description of each dataset we used in your experiments:
\paragraph{POPE.} 
POPE \cite{LiDZWZW23} is designed to detect object hallucinations using 9,000 image-question pairs. The questions are about the presence of objects (e.g., "Is there a person in the image?") and are categorized
into three sampling settings based on the selection method of nonexistent objects: random, popular,
and adversarial. 
In the random setting, nonexistent objects are chosen randomly. 
In the popular setting, objects are selected from a pool of those most frequently occurring, whereas in the adversarial setting, objects that often co-occur but are absent in the image are chosen. 
In our experiment, we focus exclusively on the adversarial setting, as it is the most challenging setting than the others and better represents the complex hallucination aspects of real-world adaptation. The evaluation metrics
used are accuracy, precision, recall, and F1-score

\paragraph{PhD.} Prompted Hallucination Dataset (PhD, \cite{abs-2403-11116}) is a carefully designed benchmark, aiming to
validate different MLLMs in intrinsic vision-language hallucination (IVL-Hallu) tasks.
Specifically, this dataset consists of four types of LVLM hallucinations: object, attribute, multi-modal conflicting, and counter-commonsense hallucinations. 
The simple Accuracy metric is used to evaluate the performance, considering that most questions result in answers that are either `yes' or `no' or are open-ended, consisting of only a few words.

\paragraph{AMBER.} An LLM-free Multi-dimensional Benchmark (AMBER, \cite{abs-2311-07397}) is a comprehensive coverage of evaluations for various types of hallucination, including those of object existence, attribute, and relation.

\begin{table*}[!th]
\centering
\fontsize{9}{12}\selectfont
\setlength{\tabcolsep}{1.9mm}
\caption{Performance comparison of models on generative and discriminative tasks in the AMBER dataset \cite{abs-2311-07397}. *: scores are copied from \cite{abs-2311-07397}.}
\label{tab:AMBER}
\begin{tabular}{@{}lccccccccc@{}}
\toprule
\multirow{2}{*}{\textbf{Model}}& \multicolumn{4}{c}{\textbf{Generative Task}} & \multicolumn{4}{c}{\textbf{Discriminative Task}} & \multirow{2}{*}{\textbf{AMBER Score}} \\
\cmidrule(lr){2-5} \cmidrule(lr){6-9} 
& \textbf{CHAIR}$\downarrow$ & \textbf{Cover}$\uparrow$ & \textbf{Hal}$\downarrow$ & \textbf{Cog}$\downarrow$ & \textbf{Acc.} & \textbf{Pre.} & \textbf{Rec.} & \textbf{F1} &  \\
\midrule
 \midrule
\bf LLaVA-1.5$^*$ & 7.8 & 51.0 & 36.4 & 4.2 & 72.0 & 93.2 & 62.4 & 74.7 & 83.5 \\
\rowcolor{lightblue}\quad + ours & 5.6 & 61.2 & 25.83 & 34.69 & 80.6  & 94.2& 71.8 & 81.5  & 87.9 (\textcolor{red}{+4.4}) \\ 
 \midrule
\bf Qwen-VL$^*$ & 5.5 & 49.4 & 23.6 & 1.9 & 81.2 & 90.8 & 79.7 & 84.9 & 89.7 \\
\rowcolor{lightblue}\quad + ours & 4.2 & 68.3 & 25.83 & 34.69 & 83.6  & 92.7 & 88.9 & 90.7 & 93.2 (\textcolor{red}{+3.5})\\ 
 \midrule
\bf GPT-4V$^*$ & 4.6 & 67.1 & 30.7 & 2.6 & 83.4 & 84.9 & 90.1 & 87.4 & 91.4 \\
\rowcolor{lightblue}\quad + ours & 3.4 & 78.9 & 25.83 & 34.69 & 89.9  & 94.4& 91.1& 92.7  & 94.5 (\textcolor{red}{+3.1})\\ 
\bottomrule
\end{tabular}
\end{table*}

\paragraph{VQAv2-IDK.} VQAv2-IDK \cite{abs-2402-09717} is proposed to evaluate the `I Know (IK)' hallucination where MLLMs generally tend to respond to the user’s question plausibly, even if the desired answer is `I don’t know'. 
VQAv2-IDK is a subset of VQAv2 \cite{GoyalKASBP19} comprising unanswerable image-question pairs as determined by human annotators. In this benchmark, `I Know (IK)' hallucination has been further categorized into four types:
\begin{itemize}
    \item Unanswerable: no one can know.
    \item Don’t know: humans may not know, but robots might.
    \item False questions: refers to non-existing.
    \item Not sure: ambiguous to answer
\end{itemize}

\paragraph{Hal-Eval.}
Hal-Eval \cite{abs-2402-15721} includes both discriminative evaluation and generative evaluation and is capable of effectively evaluating different types of hallucinations: object hallucination, attribute hallucination, relation hallucination and event hallucination.

\paragraph{WHOOPS!.}
WHOOPS! \cite{GuettaBHSES023} dataset focuses on challenging AI models to reason about commonsense and compositionally. 
There are a broader variety of tasks, in particular: (i) Explanation Generation, (ii) Image Captioning, (iii) Cross-Modal Matching, and (iv) Compositional VQA.
We evaluate our method on the Compositional VQA and Explanation Generation tasks.

\subsection{Detailed Baselines}

Here, we give a detailed description of the baselines we compared in your experiments:

\paragraph{Woodpecker.} Woodpecker \cite{abs-2310-16045} is introduced in a post-remedy manner to pick out and correct hallucinations from the generated text.
Specifically, it consists of five stages: key concept extraction, question formulation, visual knowledge validation, visual claim generation, and hallucination correction.

\paragraph{LURE.} LVLM Hallucination Revisor (LURE, \cite{abs-2310-00754}) is designed to train a post-hoc rectify object hallucination in LVLMs by reconstructing less hallucinatory descriptions.
LURE is grounded in a rigorous statistical analysis of the key factors underlying object hallucination, including co-occurrence (the frequent appearance of certain objects alongside others in images), uncertainty (objects with higher uncertainty during LVLM decoding), and object position (hallucination often appears in the later part of the generated text).

\paragraph{LogicCheckGPT.} LogicCheckGPT \cite{abs-2402-11622} proposes to adopt the logical closed loop in the context of object hallucination alleviation in LVLMs in a training-free manner. 
Specifically, the first stage of the designed framework involves inquiring about attributes based
on objects, followed by inquiring about objects based on attributes and whether their responses can form a logically closed loop, serves as an indicator of object hallucination.

\paragraph{DVP.} This work \cite{abs-2403-13513} tries to embed counterfactual thinking through specific keywords to improve the reliability and trustworthiness of LMMs' responses. 
Technically, they introduce a Dual-modality Verification Process (DVP) to ensure the precision of selecting counterfactual keywords effectively implants counterfactual thoughts.

\begin{table}[!t]
    \centering
    \fontsize{8}{9}\selectfont
    \setlength{\tabcolsep}{2.8mm}
    \begin{tabular}{@{}lccc@{}}
    \toprule
    \multirow{1}{*}{Model} & \textbf{POPE} & \textbf{PhD-SA} & \textbf{ Hal-Eval} \\  
    \midrule
    \rowcolor{lightblue} LLaVA-1.5(Ours) & 91.5 & 46.5 & 52.6   \\
    \midrule
    \quad w/ object+Attribute & 86.1 & 43.9 & 46.2   \\
    \quad w/ object+Relation& 89.0 & 45.5 & 49.5   \\
    \bottomrule
    \end{tabular}
    \vspace{-2mm}
    \caption{The effect of using partial scene graph.}
    \label{tab:psgg}
    \vspace{-2mm}
\end{table}

\begin{figure}[!t]
    \centering
    \includegraphics[width=0.85\columnwidth]{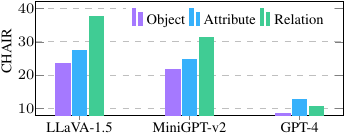}
    \vspace{-2mm}
    \caption{The hallucination rate of three MLLMs in terms of object, attribute, and relation. }
    \label{fig:hallrate}
    \vspace{-4mm}
\end{figure}

\section{Extend Results}
\label{extend_results}

\subsection{Extended Evaluation on AMBER dataset.}
We also conduct experiments on the AMBER dataset, which focuses on vision hallucinations, specifically in objects, attributes, and relationships.
As can be seen in Table \ref{tab:AMBER}, MLLMs incorporated with our methods gain a significant task improvement, indicating the effectiveness of the proposed reasoning framework.

\subsection{Extended Experiments on Hallucination Rate}

Here, through the verification step, we quantify the hallucination rate for each element in the SG. 
As illustrated in Figure \ref{fig:hallrate}, the MLLM is more prone to hallucinating relationships compared to objects and attributes.

\begin{figure*}[!t]
\centering
\includegraphics[width=1.0\textwidth]{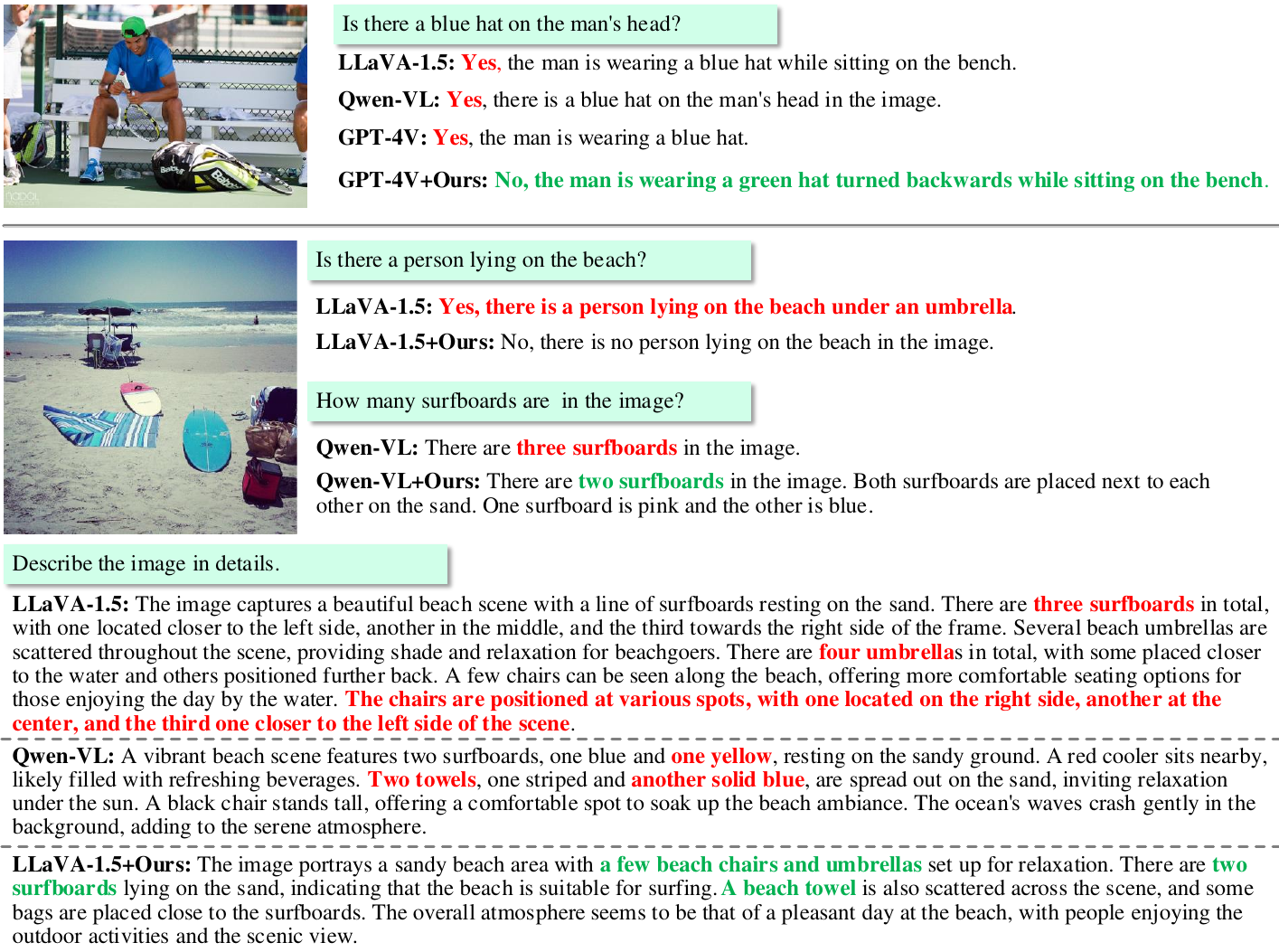}
\caption{
Examples where baseline MLLMs fail but our method succeeds.
}
\vspace{-3mm}
\label{fig:case}
\end{figure*}

\subsection{Extended Experiments on the Impact of Partial Scene Graph}
We explore the impact of utilizing the partial scene graph on task performance. 
The experimental findings in Table \ref{tab:psgg} suggest that employing a partial SG invariably leads to a decline in performance.

\begin{figure*}[!t]
\centering
\includegraphics[width=1.0\textwidth]{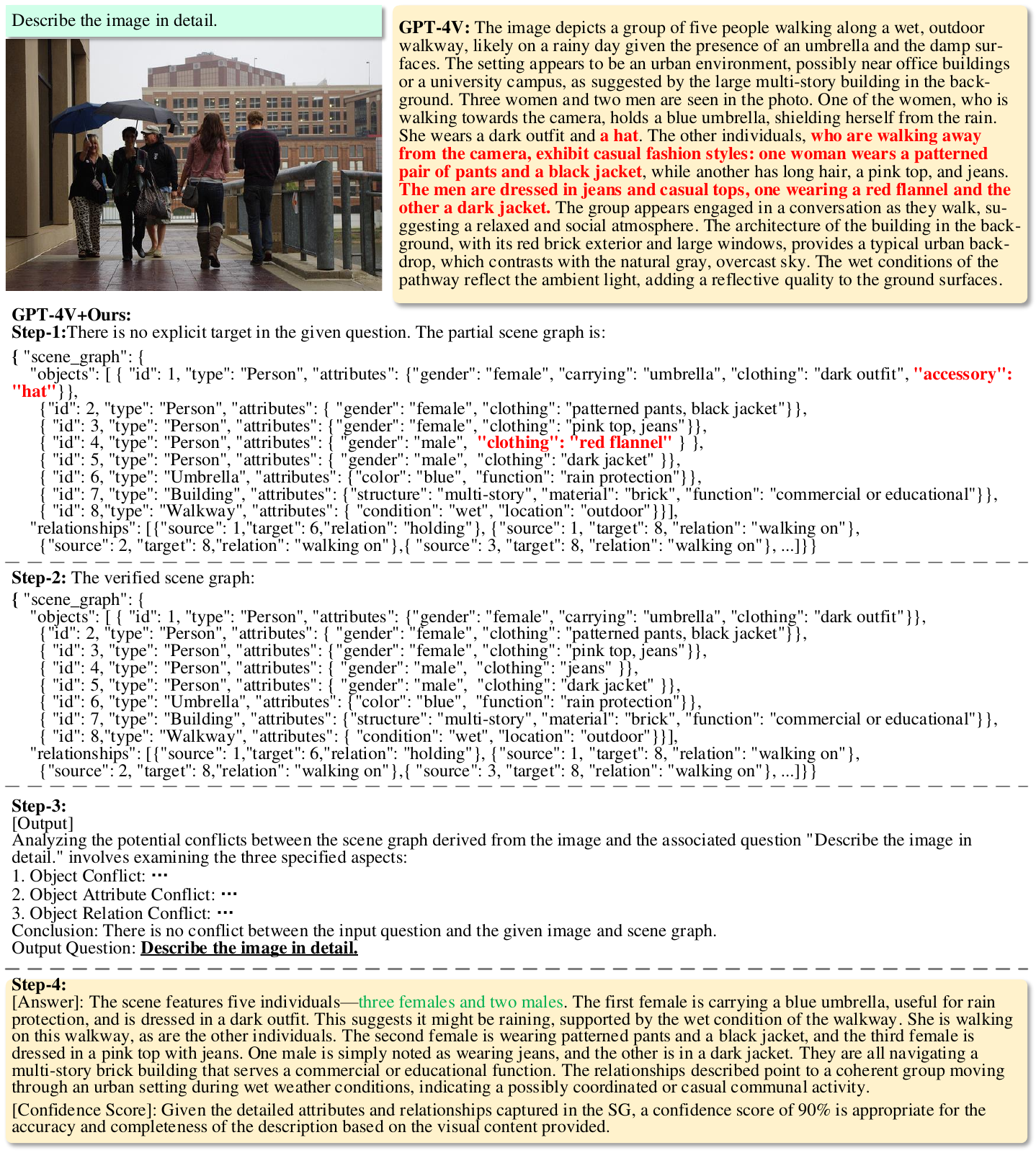}
\caption{
The illustration of examples in which step-5\&6 are not needed to answer the question.
}
\vspace{-3mm}
\label{fig:stepoutpuwithoutcommonse}
\end{figure*}

\subsection{Extended Case Study.}
\label{app:ext_case_study}

Here, we present additional examples in Figure \ref{fig:case} demonstrating how integrating our method with various MLLMs signification reduces hallucination when dealing with different types of questions. 
Furthermore, in Figures \ref{fig:stepoutpuwithoutcommonse}, we provide detailed outputs for each step involved in responding to the given questions.

\end{document}